\documentclass[runningheads]{llncs}

\usepackage[mobile]{eccv}

\usepackage{eccvabbrv}

\usepackage{graphicx}
\usepackage{booktabs}

\usepackage[accsupp]{axessibility}  

\usepackage{adjustbox}
\usepackage{multirow}

\usepackage{algorithm}
\usepackage{algorithmic}
\usepackage{amsmath}
\usepackage{amsfonts}
\usepackage{mathtools}
\usepackage{bm}
\usepackage{xcolor}
\usepackage[table]{xcolor}

\usepackage{hyperref}

\usepackage{orcidlink}

\begin{document}

\def\method{STAR}
\def\methodfull{Spanning Tree Autoregressive}

\title{Spanning Tree Autoregressive Visual Generation}



\author{
    Sangkyu Lee\inst{1,2}$^ ,$\thanks{Work mainly done during internship at LG AI Research.}\orcidlink{0009-0000-8009-1990} \and Changho Lee\inst{3}\orcidlink{0009-0007-7112-910X} \and Janghoon Han\inst{3}\orcidlink{0009-0007-3937-1764} \and Hosung Song\inst{3}\orcidlink{0009-0007-4044-4481} \and \\
    Tackgeun You\inst{2}\orcidlink{0009-0009-4253-9139} \and Hwasup Lim\inst{2}\orcidlink{0000-0003-2957-668X} \and Stanley Jungkyu Choi\inst{3}\orcidlink{0009-0001-0297-4269} \and \\
    Honglak Lee\inst{3,4}\orcidlink{0000-0002-4109-327X} \and Youngjae Yu\inst{5}$^{,\dagger}$\orcidlink{0000-0002-5867-0782}
}

\authorrunning{Lee et al.}


\institute{
    Yonsei University, Seoul, Korea \and 
    Korea Institute of Science and Technology, Seoul, Korea \and
    LG AI Research, Seoul, Korea \and
    University of Michigan, Ann Arbor, USA \and
    Seoul National University, Seoul, Korea
}

\maketitle
\begingroup
\renewcommand{\thefootnote}{$\dagger$}
\footnotetext{Corresponding author}
\endgroup

\begin{abstract}

We present \methodfull{} (\method{}) modeling, which can incorporate prior knowledge of images, such as center bias and locality, to maintain sampling performance while also providing sufficiently flexible sequence orders to accommodate image editing at inference time. Approaches that expose conventional autoregressive (AR) models in visual generation to arbitrary sequence orders via random permutation suffer from degraded sampling performance or compromise the flexibility in sequence order choice at inference time. Instead, \method{} utilizes traversal orders of uniform spanning trees in a lattice defined by the positions of image patches. Traversal orders are obtained via breadth-first search, allowing us to efficiently construct a spanning tree via rejection sampling whose traversal order ensures that the connected partial observation of the image appears as a prefix for native image inpainting support. Through the tailored yet structured sequence order randomization strategy, \method{} preserves the capability of postfix completion while maintaining sampling performance, without any significant changes to the model architecture widely adopted in language AR modeling.\begingroup\renewcommand{\thefootnote}{$\ddagger$}\footnote[1]{The code is available at \url{https://github.com/oddqueue/star}.}\endgroup

\end{abstract}
\section{Introduction}
\label{sec:introduction}

The scalability of the \textit{next-token prediction} pretraining scheme with autoregressive (AR) models built on a Decoder-only Transformer architecture~\cite{radford2019language, kaplan2020scaling, henighan2020scaling} has yielded remarkable results in language generation, represented as large language models~\cite{radford2019language, touvron2023llama}.
Meanwhile, the success of the Decoder-only Transformer has led to the transition of AR modeling for visual generation~\cite{van2016pixel, van2017neural} to the \textit{next-patch prediction} scheme, which typically predicts the next token representing a single patch by the predefined \textit{raster-scan} order. 
Similar to the language generation, AR modeling for visual generation through the next-token prediction scheme also exhibits efficient scaling behavior~\cite{liu2024elucidating, kilian2024computational, el2024scalable}, highlighting the competitive advantage of AR visual modeling compared to other methods for visual generation~\cite{chang2022maskgit, ho2020denoising, bao2023all, peebles2023scalable, li2025autoregressive}.
Furthermore, the architecture that seamlessly integrates with language generation serves as a cornerstone of early-fusion multimodal models~\cite{team2024chameleon} for native support for visual and language generation.

\begin{figure}[t]
  \centering
  \includegraphics[width=\textwidth]{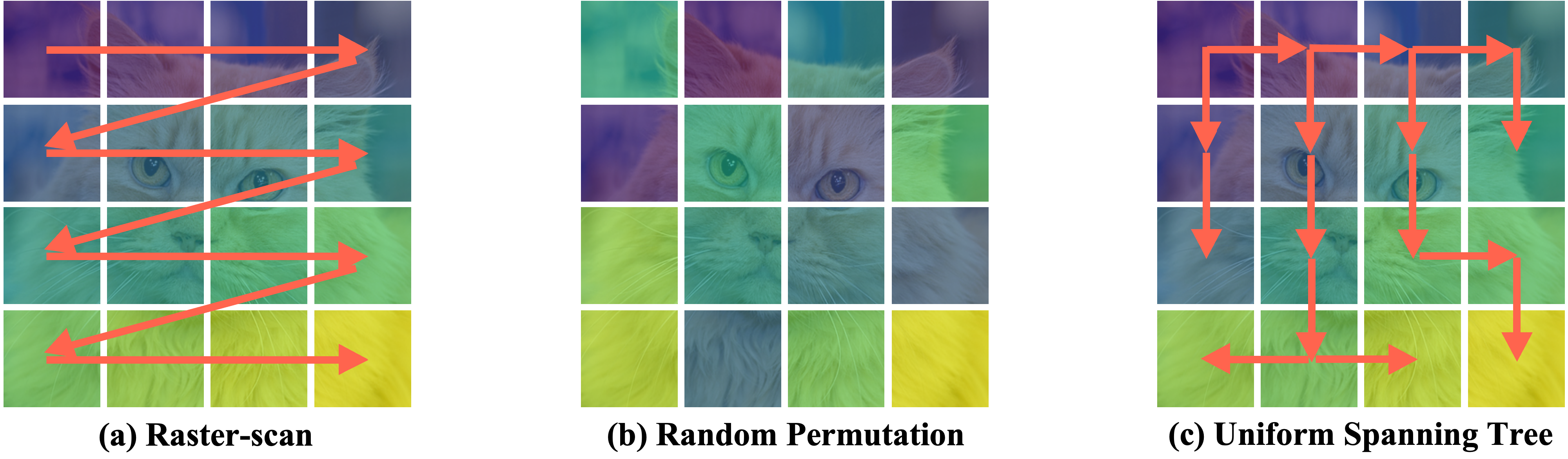}
  
  \vspace{-6pt}
  \caption{(a) Conventional AR visual generation models follow a fixed raster-scan order, limiting the flexibility of inference sequence order. (b) Random permutation offers flexibility in the inference sequence order, but does not reflect prior knowledge, such as center bias and locality. (c) \method{} adopts the traversal order of the uniform spanning tree, thereby structurally maintaining prior knowledge and flexibility at inference time.}
  
  \vspace{-18pt}
  \label{fig:teaser}
\end{figure}

However, conventional AR visual modeling restricts the 2D nature of images to a specific unidirectional dependency defined by the fixed sequence order.
To address the unidirectional problem of AR models in visual generation, recent approaches~\cite{pang2025randar, yu2025randomized, li2025autoregressive} adopt the \textit{random permutation}~\cite{yang2019xlnet, chen2020generative, pannatier2024sigma} to shuffle the sequence order of tokens and provide next-patch positions to be predicted as additional conditions to expose AR models to the diverse sequence order during training.
Nevertheless, AR models trained with arbitrary sequence orders tend to exhibit degraded performance compared to conventional AR models that follow a fixed raster-scan order, despite having flexibility in choosing inference sequence orders required for native region-level modification capabilities to achieve image editing while maintaining the Decoder-only Transformer architecture.

In this paper, we present \textbf{\methodfull{} (\method{})} modeling, a structured sequence order randomization strategy based on the \textit{uniform spanning tree} for AR visual generation models by revisiting the characteristic of raster-scan order and random permutation, as described in \Cref{fig:teaser}.
Training solely in raster-scan order forces the choice of sequence order at inference time.
However, it guarantees continuity in the 2D lattice between the tokens in the prefix and postfix. Also, it starts from the corner where salient objects are unlikely to appear.
On the other hand, random permutation does not guarantee this property of continuity and starting from the corner, but it does guarantee diverse sequence orders due to randomness.
This is the motivation of our approach, which traverses uniform spanning trees, as a representative structure that satisfies the advantages of both, reflecting \textit{prior knowledge} of images, center bias~\cite{szabo2022mitigating, borji2015reconciling} and locality~\cite{huang1999statistics, besnier2025halton}, yet preserving \textit{randomness} for flexibility.

Specifically, we consider the traversal order obtained by breadth-first search (BFS) of the uniform sampling tree, with a random corner as the root node, as the sequence order.
By choosing the BFS traversal order, we can efficiently sample a spanning tree that can handle image inpainting as a postfix completion via rejection sampling at inference time, when the given partial observation of the image is connected in the lattice, unlike the conventional raster-scan order AR visual generation model. 
Furthermore, \method{} requires minimal changes to the conventional Decoder-only Transformer, only adding additional features to provide positional information that condition the next-token position~\cite{pannatier2024sigma, yu2025randomized}.
This feature is shared with the random permutation strategy, but \method{} preserves image generation performance while still supporting image inpainting.

The contribution of our work, \method{}, can be summarized as threefold:
\begin{itemize}
    \item We propose a structured sequence order randomization strategy for AR visual generation based on the BFS traversal of the uniform sampling tree, ensuring both flexibility of sequence order and sampling performance.
    \item We show that the BFS traversal order of uniform spanning trees is sufficient to achieve the benefits of random permutation for diverse sequence orders in AR visual generation models without increasing training complexity.
    \item We introduce a new building block that supports region-level modification capabilities for image editing with minimal changes that do not sacrifice performance, a challenge in conventional AR visual generative models.
\end{itemize}

\section{Related Work}
\label{sec:related_work}

\subsection{Unidirectional Autoregressive Visual Generation}
Decoder-only Transformer proposed by GPT-2~\cite{radford2019language}, which adopts the causal attention mask, has become the de facto standard architecture for AR language modeling~\cite{brown2020language, touvron2023llama} due to its advantageous feature of scaling behavior~\cite{kaplan2020scaling, henighan2020scaling}.
The success of the Decoder-only Transformer, often represented by next-token prediction, has enriched the research direction of image generation based on AR modeling~\cite{van2016pixel, van2017neural} to next-token prediction.
A sequence of research for next-token prediction for image generation~\cite{esser2021taming, sun2024autoregressive, yu2021vector} achieves AR visual modeling by training models to predict patch representations of images, typically arranged in a predefined, unidirectional order.
Here, a series of works~\cite{esser2021taming, el2024scalable, yu2025randomized} commonly observe that raster-scan order outperforms variants of sequence order, such as space-filling curves, for the predefined unidirectional order, and raster-scan is widely adopted as the standard sequence order for AR visual modeling.
Sharing the architecture across data modalities facilitates expansion into early-fusion multimodal models~\cite{team2024chameleon}, but unidirectional modeling of images imposes limitations on image editing due to their 2D structure.
Inspired by this aspect, we aim to enable flexible downstream task capabilities for AR visual generation models without significant architectural changes from the Decoder-only Transformer.

\subsection{Bidirectional Modeling for Visual Generation}
A line of research has been conducted to address the bidirectional dependency between image patches in visual generation.
One research direction is to directly introduce bidirectional transformers for masked token prediction~\cite{chang2022maskgit}, diffusion models~\cite{ho2020denoising, bao2023all, peebles2023scalable}, or to mimic AR models with diffusion loss~\cite{li2025autoregressive}.
However, they often show less efficiency than the AR models in scaling behavior~\cite{liu2024elucidating, kilian2024computational}.
Another approach decouples spatial prediction in AR visual modeling and introduces a dedicated transformer architecture~\cite{lee2022autoregressive, tian2024visual, ren2024flowar, liu2025detailflow, roheda2026cart}, but this can limit the simplicity of extension to early-fusion multimodal models.
The other approach exposes AR models by applying random permutation to the sequence order~\cite{chen2020generative} and providing the conditions for next-token position through position instruction tokens~\cite{pang2025randar}, additional positional embeddings~\cite{yu2025randomized, pannatier2024sigma}, or multi-stream attention mechanisms~\cite{yang2019xlnet, li2025autoregressive}, but commonly trade either sampling performance or flexibility in sequence order at inference time.
As an extension of the last approach, we investigate how to extend the trade-off frontier between performance and flexibility while preserving the Decoder-only Transformer architecture.

\section{Method}
\label{sec:method}

\begin{figure}[t]
  \centering
  \includegraphics[width=\textwidth]{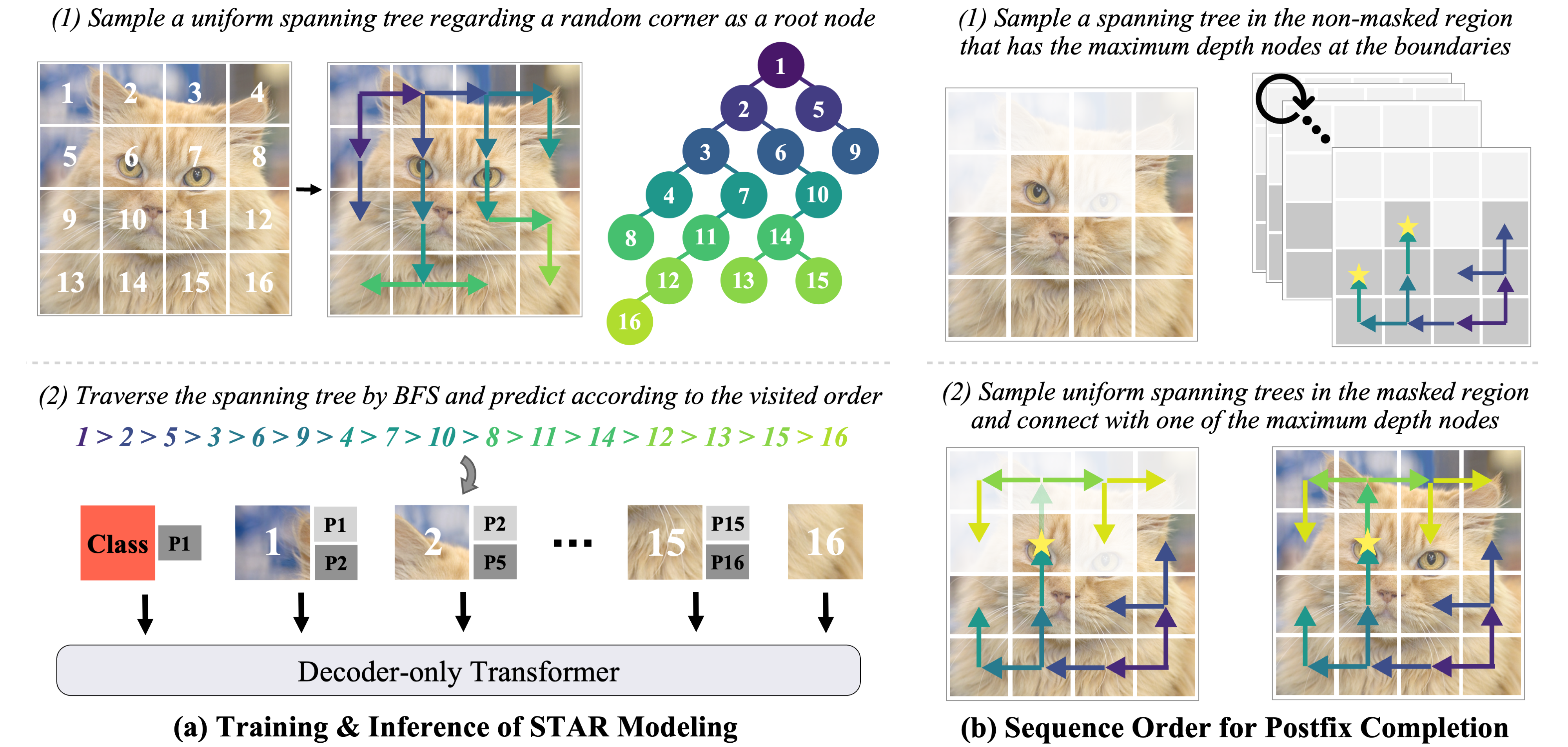}
  
  \vspace{-6pt}
  \caption{Overview of \method{}. (a) We perform training and inference in the sequence order of a BFS traversal for uniform spanning trees that start at the corners, additionally conditioning on the next-token positions. (b) We perform rejection sampling of the spanning tree with the maximum depth at the boundaries of the non-masked region, allowing postfix completion of partial observations to achieve region-level modification.}
  
  \vspace{-16pt}
  \label{fig:overview}
\end{figure}

In this section, we describe our approach, \textbf{\methodfull{} (\method{})} modeling, an autoregressive visual modeling based on the traversal order of the uniform spanning tree to impose structurally randomized sequence orders preserving both of image generation performance and flexibility in choice of sequence orders to accommodate image editing, as described in \Cref{fig:overview}.

\subsection{From Fixed To Randomized Sequence Orders}
As generative models for visual generation, AR models approximate the marginal distribution of the image $x$ by factorizing it into conditional distributions trained with maximum likelihood estimation.
Here, the conventional approach treats an image as a collection of patch token representations $\{ x_{i,j} \}$ of size $h \times w$. 
These 2D-structured tokens are linearized to the 1D sequence order $\tau \coloneqq (\tau_1, \cdots \tau_N): \mathbb{Z}_h \times \mathbb{Z}_w \rightarrow \mathbb{Z}_N$, which serves as a prediction order for AR models where $N = h \times w$.
Formally, AR models $p_\theta$ for visual generation are trained as follows:
\begin{equation}
  \underset{\theta}{\text{argmax}} \,\, p_\theta(x) = p_\theta(x_{\tau_0}) \prod_{i=1}^{N} p_\theta(x_{\tau_i}|x_{\tau_1}, x_{\tau_2}, \cdots x_{\tau_{i-1}}) = \prod_{i=0}^{N} p_\theta(x_{\tau_i}|x_{\tau_{<i}}).
  \label{eq:ar}
\end{equation}

The choice of sequence order $\tau$ is not a straightforward feature in visual generation compared to natural language generation.
It is challenging to impose an adequate sequence order on images because of their 2D structure, unlike natural languages, which exhibit a relatively straightforward left-to-right sequence order.
The empirically practical approach is to adopt the well-known sequence order, specifically the raster-scan order, as it often demonstrates better performance compared to other sequence orders~\cite{esser2021taming, el2024scalable, yu2025randomized}. 
However, adopting a single fixed sequence order imposes a structural limitation on the AR models: it exposes the image to the model only in a single, unidirectional sequence. 
In other words, it prevents conditioning on tokens that do not appear as prefixes in the predefined sequence order, unlike bidirectional modeling for visual generation~\cite{chang2022maskgit, peebles2023scalable}.

In this context, some approaches uniformly sample multiple sequence orders from random permutations $S_N$ during training, rather than fixing to a single order~\cite{chen2020generative, pang2025randar, li2024autoregressive, yu2025randomized, li2025autoregressive}.
The common goal is to enable AR models to be trained on randomly ordered contexts as a bidirectional modeling in \textit{expectation}:
\begin{equation}
  \underset{\theta}{\text{argmax}} \,\, p_\theta(x) = \mathbb{E}_{\tau \sim S_N} \Big[ \prod_{i=0}^{N} p_\theta(x_{\tau_i}|x_{\tau_{<i}}) \Big].
  \label{eq:random_ar}
\end{equation}

This approach has been investigated under the name of \textit{permutation} AR modeling~\cite{yang2019xlnet, pannatier2024sigma} beyond visual generation.
Meanwhile, it is known that this loss objective is theoretically equivalent to the loss objective for masked token prediction~\cite{uria2016neural, ou2024your}, and the difference between the two approaches is whether they use the Decoder-only Transformer as the model architecture or not ~\cite{xue2025any}.
Still, this flexibility in sequence order adds a key downstream feature for visual generation: native support of region-level modification required for image editing.
Given the 2D position indices of non-masked patches $\hat{I} \times \hat{J}$, we can achieve image inpainting by simply finding a sequence order $\hat{\tau} \coloneqq (\tau_\text{pre}, \tau_\text{post})$ that ensures the partial observation of the image appears as a prefix for completing the postfix:
\begin{equation}
    \tau_\text{pre}: \hat{I} \times \hat{J} \rightarrow \mathbb{Z}_{|\hat{I} \times \hat{J}|},\,\tau_\text{post}: (\mathbb{Z}_h \times \mathbb{Z}_w) \setminus (\hat{I} \times \hat{J}) \rightarrow |\hat{I} \times \hat{J}| + \mathbb{Z}_{|(\mathbb{Z}_h \times \mathbb{Z}_w) \setminus (\hat{I} \times \hat{J})|}.
  \label{eq:inpainting}
\end{equation}

However, this permutation AR modeling often converges more slowly than conventional AR modeling that uses a single predefined sequence known to be advantageous for the intrinsic structure~\cite{xue2025any}.
In theory, if we could perfectly model the data distribution, the model entropy $\mathcal{H}_{\theta}(x)$ should converge to the data entropy $\mathcal{H}(x)$ for any permutation because of the chain rule for entropy (i.e $\mathcal{H}(x) \approx \sum_{i=0}^N \mathcal{H}_\theta(x_{\tau_i}|x_{\tau_{<i}})\,\, \forall \tau \in S_N$), but empirically this does not happens because of the different convergence speed across sequence orders.
The existence of advantageous sequence orders in convergence speed, such as the raster-scan order, has been repeatedly observed in previous work~\cite{esser2021taming, el2024scalable, yu2025randomized}.
Nevertheless, permutation AR modeling is forced to maximize the likelihood over all sequence orders rather than the advantageous ones, leading to inferior performance compared to the conventional AR model within the same computation budget.

\subsection{Uniform Spanning Tree for Autoregressive Modeling}
\label{sec:method_analysis}

The inherent nature of permutation AR modeling underscores the need for an approach that improves convergence speed while retaining the advantage of native image inpainting capability.
Nonetheless, prior work leverages permutation AR modeling only as a pretext task for conventional AR models~\cite{yu2025randomized} or focuses on parallel inference~\cite{pang2025randar, li2025autoregressive} as similarly investigated in masked token prediction~\cite{chang2022maskgit}.
It leads to increased training complexity due to additional hyperparameter search while sacrificing native image inpainting capability, or directly results in inferior sampling performance compared to AR models using the raster-scan order.

The native image inpainting capability can be achieved if the model supports random sequence orders that are sufficient to find a specific sequence order that makes a given partial observation appear in the prefix and a masked region appear in the postfix.
However, random permutation exposes all $N!$ possible sequence orders beyond the tighter sets that support native image inpainting, including non-advantageous orders that can affect convergence speed.
Therefore, we need to narrow the sample space of sequence orders by identifying the common properties underlying well-known predefined sequence orders, such as raster-scan order and space-filling curves (e.g., Hilbert and Morton curves).

In this respect, we propose an approach that uses the \textit{uniform spanning tree} on a lattice corresponding to the token position relationships in images to structurally reflect prior knowledge about the image in the sequence order randomization strategy.
Formally, we regard the 2D token position index $(i, j)$ as a vertex and each adjacent vertex $u, v \in V$ is connected by edge $(u, v) \in E$ as a regular taxicab lattice $G \coloneqq (V, E)$.
We uniformly sample a spanning tree $T \sim \mathcal{T}(G, r)$, which is a subgraph of $G$ that forms a tree that includes all vertices of $G$ while considering a randomly chosen corner vertex as the root node $r$.
Then, we train the AR model $p_\theta$ by maximum likelihood estimation with additional conditioning on the next-token position according to the traversal order $\tau_s$:
\begin{equation}
  \underset{\theta}{\text{argmax}} \,\, \mathbb{E}_{\tau \sim \tau_{s} \big({\mathcal{T}(G, r)}\big)} \Big[\prod_{i=0}^{N} p_\theta(x_{\tau_i}|x_{\tau_{<i}}) \Big].
  \label{eq:star}
\end{equation}

\begin{figure}[t]
  \centering
      \begin{subfigure}{0.62\columnwidth}
        \includegraphics[width=\columnwidth]{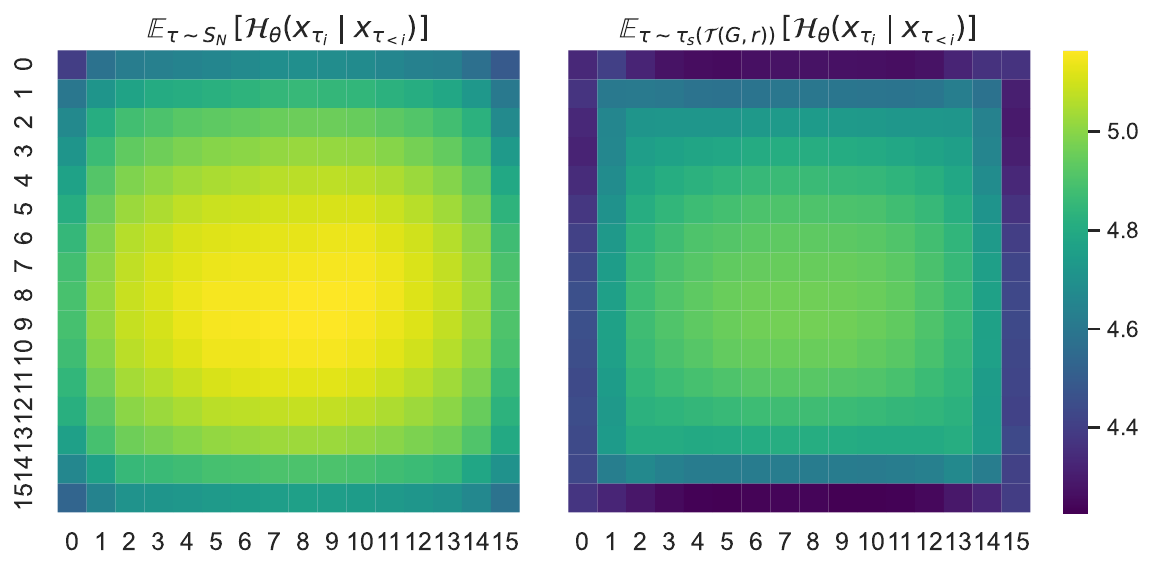}
        \caption{}
        \label{fig:entropy}
      \end{subfigure}
      \hfill
      \begin{subfigure}{0.37\columnwidth}
        \includegraphics[width=\columnwidth]{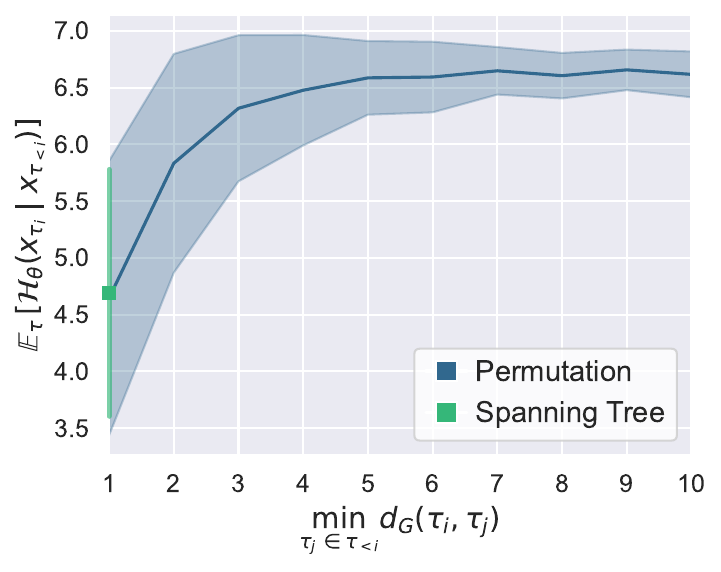}
        \caption{}
        \label{fig:conditional_entropy}
      \end{subfigure}
      
  \vspace{-6pt}
  \caption{(a) Conditional model entropy according to token position. Higher token entropy in central regions indicates that token position affects the differences in prediction difficulty. (b) Conditional model entropy according to the minimum Manhattan distance from the prefix token positions. Tokens located at closer distances tend to exhibit lower entropy, suggesting that adjacent tokens are easier to predict than others.}
  
  \vspace{-14pt}
\end{figure}

This stems from the conjecture that the advantage of predefined sequences is derived from \textit{prior knowledge} of the image, such as \textit{center bias}~\cite{szabo2022mitigating, borji2015reconciling} and \textit{locality}~\cite{huang1999statistics, besnier2025halton}.
To verify whether similar phenomena persist in token representations of images, we measure and analyze conditional model entropy $\mathcal{H}_{\theta}(x_{\tau_i} \mid x_{\tau_{<i}})$ on validation images of ImageNet-1k~\cite{deng2009imagenet} by sampling 30 sequence orders per single image for each AR model trained with random permutation and uniform spanning tree.
\Cref{fig:entropy} and \Cref{fig:conditional_entropy} demonstrate that prior knowledge of the image is maintained in token representations; AR models show difficulty in predicting token positions moving away from the already predicted tokens or moving towards the central region, which is likely to include complex salient objects.
In particular, the fact that the conditional model entropy increases rapidly with increasing distance from adjacent tokens suggests that partial observations cannot serve as meaningful context unless they are connected to the token being predicted, a similar observation to that found in masked token prediction~\cite{besnier2025halton}.

However, the sequence order of traversing the uniform spanning tree from the corners naturally moves towards the central region from the corner while predicting adjacent tokens to the prefix.
Furthermore, it is known that the number of spanning trees asymptotically follows $\exp(N \cdot z_G)$ as $N \rightarrow \infty $ where $z_G$ is a constant defined by the lattice type, $z_G \approx 1.166$ for our regular taxicab lattice~\cite{shrock2000spanningtrees}.
Therefore, the number of spanning trees exhibits exponential growth with $N$, a distinctly smaller sample space compared to the $N!$ sample space of random permutation.
Still, the uniform spanning tree can be efficiently sampled based on the loop-erased random walk, also known as Wilson's algorithm, requiring $O(N \log N)$ time for sampling a uniform spanning tree in the 2D lattice~\cite{wilson1996generating}.

\subsection{Breadth-first Search Traversal for Postfix Completion}
\begin{algorithm}[t]
    \caption{Sampling Sequence Orders for Postfix Completion}
    \label{alg:postfix_completion}
    \begin{algorithmic}[1]
        \REQUIRE lattice $G \coloneqq (V, E)$, mask $M \coloneqq (V_M,E_M)$, corners $R \subset V \setminus V_M$
        \ENSURE $M \subset G$, $G \setminus M$ is connected, $R \not = \emptyset$
        \STATE \texttt{\# (a) Find boundaries of the non-masked region and the farthest root}
        \STATE $G'\gets (V \setminus V_M,\, \{(u, v) \in E \mid u, v \notin V_M \})$
        \STATE $B \gets \bigcup_i B_i = \bigcup_i \{v\in V \setminus V_{M_i} \mid \exists\,u\in V_{M_i} \,\, \text{s.t. } (u, v) \in E \}$
        \STATE $r_{G'} \gets \arg \max_{r \in R} \text{min}_{B_i \subset B} \frac{1}{|B_i|}\sum_{v \in B_i}d_{G'}(v, r) $
        \STATE \texttt{\# (b) Sample the spanning tree in the non-masked region by rejection}
        \REPEAT
            \STATE $T_{G'} \gets (V_{G'}, E_{T_{G'}}) \sim \mathcal{T}(G', r_{G'})$
            \STATE $V_{G'_{\text{max}}} \gets \{v \in V_{T_{G'}}\mid \arg \max_{v \in V_{G'}} \text{depth}_{T_{G'}}(v, r_{G'}) \}$
        \UNTIL {$\bigwedge_i (B_i \cap V_{G'_{\text{max}}} \ne \emptyset )$}
        \STATE \texttt{\# (c) Sample spanning trees in the masked region and connect trees}
        \STATE $E_C \gets \bigcup_i E_{C_i} =\bigcup_i \big\{ (u, v_i) \mid u \in B_i, \,\, v_i \sim \{v \in V_{M_i} \mid \exists\, u \in B_i \,\, \text{s.t. } (u, v) \in E \} \big\} $ 
        \STATE $T_M \gets (V_{M}, E_{T_M}) = \bigcup_i T_{M_i} \sim \mathcal{T}(M_i, v_i)$ 
        \STATE $T_G \gets \big( (G, E_{T_{G'}} \cup E_C \cup E_{T_M}), r_{G'} \big)$
        \RETURN $\text{sequence order } \tau_s (T_G)$
    \end{algorithmic}
\end{algorithm}

The sequence order derived from traversing uniform spanning trees structurally maintains prior knowledge of images, but this does not guarantee that we can easily find a spanning tree whose traversal order accomplishes image inpainting for a given arbitrary mask.
Unlike permutation AR modeling, which can simply sample the prefix sequence order $\tau_\text{pre}$ and the postfix sequence order $\tau_\text{post}$ by sampling permutations in each non-masked and masked region, we need to sample the spanning tree according to the traversal strategy.
Thus, we need to choose an appropriate traversal strategy that can easily sample a spanning tree that satisfies the requirement for postfix completion at inference time.

Representative approaches for traversing a tree structure include the depth-first search (DFS) and breadth-first search (BFS) algorithms.
However, under a fixed tie-breaking rule, BFS has an advantage over DFS in rejection sampling to accomplish postfix completion at inference time.
Formally, consider a typical image editing scenario with a given entire mask $M \coloneqq (V_M, E_M) = \bigcup_{i}M_i$, which is a collection of mutually disconnected masks $M_i \coloneqq (V_{M_i}, E_{M_i})$.
Assuming $M$ does not cover all corners and does not divide partial observations into disconnected graphs, the following condition must be satisfied for the spanning tree $T \sim \mathcal{T}(G, r)$ to be able to perform postfix completion in the traversal order:
\begin{equation}
  \text{depth}_T(v_{M}) \ge \text{depth}_T(v) \,\,\,\, \forall\, v \in V \setminus V_{M}, \, v_{M} \in V_{M}.
  \label{eq:condition}
\end{equation}

\begin{figure}[t]
  \centering
  
  \vspace{-11pt}
  \includegraphics[width=\textwidth]{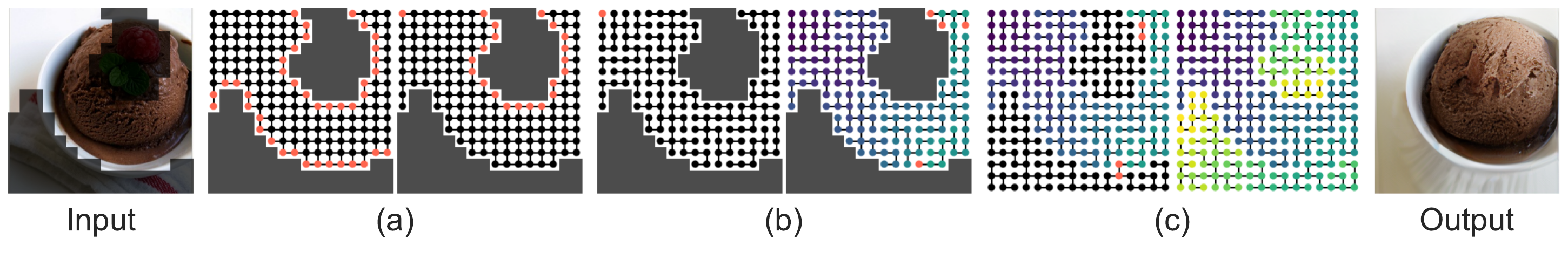}
  
  \vspace{-10pt}
  \caption{Example of sampling the sequence order for postfix completion of the given input image. We chronologically illustrate the procedures (a), (b), and (c) in \Cref{alg:postfix_completion} while highlighting $r$, $B_i$, maximum depth vertices, and spanning tree connections.}
  
  \vspace{-16pt}
  \label{fig:postfix_completion}
\end{figure}

This stems from the inherent property of BFS, where the postfix vertex depth must be greater than or equal to the prefix vertex depth.
In other words, if there is no vertex in the spanning tree whose depth is the same as the depth of a vertex in the non-masked region $V \setminus V_M$, then the depth of the vertex in the spanning tree must be shallower than that of all vertices belonging to $M$.
This is equivalent to the condition that the restriction of the spanning tree $T|_{V \setminus V_M}$ has maximum depth at each boundary $B_i$, which is the collection of vertices in ${V \setminus V_{M_i}}$ adjacent to $M_i$.
Therefore, it is sufficient to sample a spanning tree $T|_{V \setminus V_M}$ where at least one of the vertices with the maximum depth is adjacent to each $M_i$, then complete the spanning tree in $G$ by sampling spanning trees in $M_i$ and connecting to these vertices to ensure depth increments without checking tie-breaking violations.
In contrast, adopting DFS imposes a much stricter constraint even if $M$ is given as a single connected mask, since it requires exactly one vertex visited last during the traversal of $T|_{V \setminus V_M}$ to be adjacent to $M$.

Thus, in our approach, we choose the traversal order $\tau_s$ as BFS during both training and inference.
If partial observation is provided at inference time, we first sample a spanning tree in the non-masked region to enable postfix completion via rejection sampling, treating the corner farthest from the closest boundary in Manhattan distance as the root node to increase acceptance.
We select one random vertex in the masked region adjacent to the maximum depth nodes corresponding to each $B_i$ and treat it as the new root for the spanning tree in $M_i$.
Finally, we complete the spanning tree in $G$ by uniformly sampling spanning trees for $M_i$ rooted at the selected root nodes, and connecting each root node to $B_i$ using a single edge connection.
\Cref{alg:postfix_completion} and \Cref{fig:postfix_completion} demonstrate sampling sequence order for postfix completion.
Further discussion about scenarios that violate our assumption of masking is included in the supplementary material.
\section{Experiments}
\label{sec:experiments}

\subsection{Implementation Details}

\subsubsection{Dataset and Model Architecture} 
We assess the feasibility of \method{} by class-conditional generative models on ImageNet-1k~\cite{deng2009imagenet} at $256 \times 256$ resolution with minimal changes to the conventional Decoder-only Transformer, following the configurations proposed by RAR~\cite{yu2025randomized}.
We use MaskGIT-VQGAN tokenizer~\cite{chang2022maskgit}, a CNN-based VQ autoencoder trained on ImageNet-1k at $256 \times 256$ with a $ 1024$-codebook size and a $16$ downsampling rate.
We adopt the ViT~\cite{dosovitskiy2020image} architecture with additional features, including causal attention mask, QK LayerNorm~\cite{dehghani2023scaling}, class conditioning by adaLN~\cite{peebles2023scalable} with the first token of the sequence.
We choose learnable positional embeddings for conditioning of next-token positions as in previous work~\cite{yu2025randomized, pannatier2024sigma} without testing alternatives such as RoPE, which separately encodes next-token positions~\cite{heo2024rotary}, or adaLN, which regards next-token positions as token-level conditions.
We scale the model parameters into four configurations, denoted as \method{}-B, \method{}-L, \method{}-XL, and \method{}-XXL.

\subsubsection{Training and Evaluation}
\begin{table*}[t]
    \centering
    \caption{Comparison of class-conditional image generation on ImageNet-1k at $256 \times 256$. We report Fréchet Inception Distance (FID), Inception Score (IS), Precision (P), and Recall (R), with $\downarrow$ or $\uparrow$ for indicating lower or higher is better in the given metric.}
    
    \vspace{-8pt}
    \adjustbox{max width=\textwidth}{
        \setlength{\tabcolsep}{5pt}
        \begin{tabular}{cclccccc}
            \toprule
            \textbf{Tokenizer} & \textbf{Model Type} & \textbf{Model} & \textbf{\# Params} & \textbf{FID}$(\bm{\downarrow})$ & \textbf{IS}$(\bm{\uparrow})$ & \textbf{P}$(\bm{\uparrow})$ & \textbf{R}$(\bm{\uparrow})$ \\
            \midrule
            \multirow{7}{*}{VAE} & \multirow{4}{*}{Diffusion} & UViT-L/2~\cite{bao2023all} & 287M & 3.40 & 219.9 & 0.83 & 0.52 \\
            & & UViT-XL/2~\cite{bao2023all} & 501M & 2.29 & 263.9 & 0.82 & 0.57 \\
            \cmidrule(lr){3-8}
            & & DiT-L/2~\cite{peebles2023scalable} & 458M & 5.02 & 167.2 & 0.75 & 0.57 \\
            & & DiT-XL/2~\cite{peebles2023scalable} & 675M & 2.27 & 278.2 & 0.83 & 0.57 \\
            \cmidrule(lr){2-8}
            & \multirow{3}{*}{Bidirectional AR} & MAR-B~\cite{li2024autoregressive} & 208M & 2.31 & 281.7 & 0.82 & 0.57 \\
            & & MAR-L~\cite{li2024autoregressive} & 479M & 1.78 & 296.0 & 0.81 & 0.60 \\
            & & MAR-H~\cite{li2024autoregressive} & 943M & 1.55 & 303.7 & 0.81 & 0.62 \\
            \midrule
            \multirow{19}{*}{VQ} & \multirow{8}{*}{Raster-scan AR} & LlamaGen-L~\cite{sun2024autoregressive} & 343M & 3.80 & 248.3 & 0.83 & 0.51 \\
            & & LlamaGen-XL~\cite{sun2024autoregressive} & 804M & 3.39 & 227.1 & 0.81 & 0.54 \\
            & & LlamaGen-XXL~\cite{sun2024autoregressive} & 1.4B & 3.09 & 253.6 & 0.83 & 0.53 \\ 
            & & LlamaGen-3B~\cite{sun2024autoregressive} & 3.0B & 3.05 & 222.3 & 0.80 & 0.58 \\
            \cmidrule(lr){3-8}
            & & RAR-B~\cite{yu2025randomized} & 261M & 1.95 & 290.5 & 0.82 & 0.58 \\
            & & RAR-L~\cite{yu2025randomized} & 461M & 1.70 & 299.5 & 0.81 & 0.60 \\
            & & RAR-XL~\cite{yu2025randomized} & 955M & 1.50 & 306.9 & 0.80 & 0.62 \\
            & & RAR-XXL~\cite{yu2025randomized} & 1.5B & 1.48 & 326.0 & 0.80 & 0.63 \\
            \cmidrule(lr){2-8}
            & \multirow{4}{*}{Block-wise AR} & VAR-$d$16~\cite{tian2024visual} & 310M & 3.30 & 274.4 & 0.84 & 0.51 \\
            & & VAR-$d$20~\cite{tian2024visual} & 600M & 2.57 & 302.6 & 0.83 & 0.56 \\
            & & VAR-$d$24~\cite{tian2024visual} & 1.0B & 2.09 & 312.9 & 0.82 & 0.59 \\
            & & VAR-$d$30~\cite{tian2024visual} & 2.0B & 1.92 & 323.1 & 0.82 & 0.59 \\
            \cmidrule(lr){2-8}
            & \multirow{7}{*}{Randomized AR} & RandAR-L~\cite{pang2025randar} & 343M & 2.55 & 288.8 & 0.81 & 0.58 \\
            & & RandAR-XL~\cite{pang2025randar} & 775M & 2.25 & 314.2 & 0.80 & 0.60 \\
            & & RandAR-XXL~\cite{pang2025randar} & 1.4B & 2.15 & 322.0 & 0.79 & 0.62 \\
            \cmidrule(lr){3-8}
            \rowcolors{}{}{}
            & & \cellcolor{gray!20}STAR-B & \cellcolor{gray!20}261M & \cellcolor{gray!20}2.24 & \cellcolor{gray!20}295.3 & \cellcolor{gray!20}0.82 & \cellcolor{gray!20}0.57 \\
            & & \cellcolor{gray!20}STAR-L & \cellcolor{gray!20}461M & \cellcolor{gray!20}1.98 & \cellcolor{gray!20}322.0 & \cellcolor{gray!20}0.82 & \cellcolor{gray!20}0.58 \\
            & & \cellcolor{gray!20}STAR-XL & \cellcolor{gray!20}955M & \cellcolor{gray!20}1.65 & \cellcolor{gray!20}333.2 & \cellcolor{gray!20}0.80 & \cellcolor{gray!20}0.62 \\
            & & \cellcolor{gray!20}STAR-XXL & \cellcolor{gray!20}1.5B & \cellcolor{gray!20}1.55 & \cellcolor{gray!20}338.8 & \cellcolor{gray!20}0.81 & \cellcolor{gray!20}0.62 \\
            \bottomrule
        \end{tabular}
    }
    \label{tab:main}
\end{table*}

We apply the ten-crop transformation~\cite{szegedy2015going, yu2025randomized} and train for 250k steps with a batch size of 2048 ($\sim$ 400 epochs) for all configurations.
We use AdamW~\cite{loshchilov2017decoupled}, cosine learning rate scheduler with linear warmup, Wilson's algorithm~\cite{wilson1996generating} for sampling uniform spanning trees, fixed tie-breaking rule that first visits the index preceding others in raster-scan order during the BFS, class dropout with a probability of 0.1 for classifier-free guidance~\cite{ho2022classifier}, and power-cosine CFG scheduler~\cite{gao2023masked, wang2024analysis} without the top-p or top-k sampling at inference time.
We follow the evaluation protocol of ADM~\cite{dhariwal2021diffusion} and report Fréchet Inception Distance (FID)~\cite{heusel2017gans}, Inception Score (IS)~\cite{salimans2016improved}, Precision, and Recall~\cite{kynkaanniemi2019improved}.
We extend this protocol to quantitatively evaluate image inpainting performance using validation images, in which a random connected set of tokens is dropped at given masking ratios, treating the remaining tokens as partial observations.
Further implementation details are described in the supplementary material.

\begin{figure}[t]
  \centering
  
  \vspace{-13pt}
  \includegraphics[width=\textwidth]{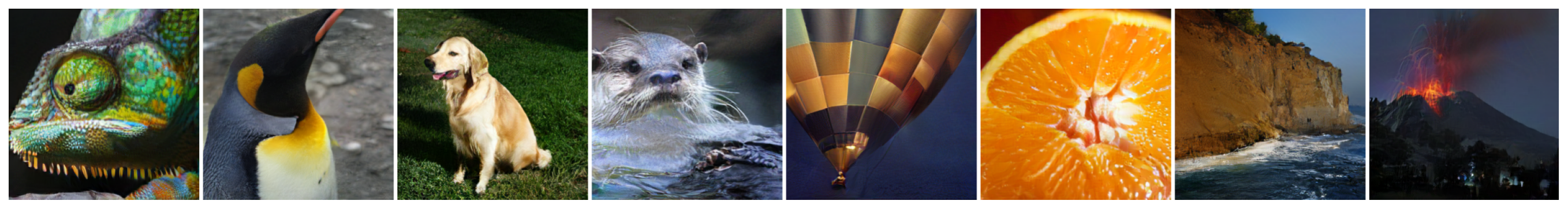}
  
  \vspace{-8pt}
  \caption{Class-conditional generation of \method{}-XXL on ImageNet-1k at $256 \times 256$.}
  
  \vspace{-16pt}
  \label{fig:generation_qualitative}
\end{figure}

\subsection{Main Results}
\subsubsection{Class-conditional Image Generation}
In \Cref{tab:main}, we compare \method{} with state-of-the-art class-conditional image generation models trained on ImageNet-1k at $256 \times 256$.
We observe that simple modification of \method{}, introducing the traversal order of the uniform spanning tree to conventional AR models, such as LlamaGen~\cite{sun2024autoregressive}, can achieve performance that is sufficiently comparable with state-of-the-art models without increasing the training complexity or architectural changes, such as the sequence order annealing strategy from random permutation to raster-scan order that requires additional hyperparameter search like  RAR~\cite{yu2025randomized} and the block-wise causal attention mask and tailored tokenizer like VAR~\cite{tian2024visual}.
In addition, \method{} outperforms RandAR~\cite{pang2025randar}, which adopts random permutation of sequence orders during training, demonstrating the effectiveness of spanning trees over random permutation in sequence order randomization.
Especially, given that MAR~\cite{li2025autoregressive} also adopts random permutation, the notable performance of \method{} trained with cross-entropy loss and a discrete VQ tokenizer suggests a promising future extension with a diffusion loss and a continuous VAE tokenizer. 
\Cref{fig:generation_qualitative} illustrates sampled results from \method{}-XXL, and more examples are included in the supplementary material.

\begin{table*}[t]
    \centering
    \caption{Further comparison of class-conditional generative models~\cite{peebles2023scalable, yu2025randomized, pang2025randar} on ImageNet-1k at $256 \times 256$ with image inpainting. We report the average scores of each metric across masking ratios from 0.1 to 0.9, in 0.1 increments, on the image inpainting.}
    
    \vspace{-8pt}
    \adjustbox{max width=\textwidth}{
        \setlength{\tabcolsep}{3pt}
        \begin{tabular}{clccccccccc}
            \toprule
            \multirow{2}{*}{\textbf{Model Type}} & \multirow{2}{*}{\textbf{Model}} & \multirow{2}{*}{\textbf{\# Params}} & \multicolumn{4}{c}{\textbf{Generation}} & \multicolumn{4}{c}{\textbf{Inpainting}} \\
            \cmidrule(lr){4-7}\cmidrule(lr){8-11}
            & & & \footnotesize\textbf{FID$(\bm{\downarrow})$} & \footnotesize\textbf{IS$(\bm{\uparrow})$} & \footnotesize\textbf{P}$(\bm{\uparrow})$ & \footnotesize\textbf{R}$(\bm{\uparrow})$ & \footnotesize\textbf{FID$(\bm{\downarrow})$} & \footnotesize\textbf{IS$(\bm{\uparrow})$} & \footnotesize\textbf{P}$(\bm{\uparrow})$ & \footnotesize\textbf{R}$(\bm{\uparrow})$ \\
            \midrule
            Diffusion & DiT-XL/2~\cite{peebles2023scalable} & 675M & 2.27 & 278.2 & 0.83 & 0.57 & 4.58 & 50.4 & 0.74 & 0.63 \\
            \midrule
            \multirow{2}{*}{Raster-scan AR} & RAR-L~\cite{yu2025randomized} & 461M & 1.70 & 299.5 & 0.81 & 0.60 & 3.56 & 86.7 & 0.79 & 0.60 \\
            & RAR-XL~\cite{yu2025randomized} & 955M & 1.50 & 306.9 & 0.80 & 0.62 & 3.40 & 88.2 & 0.79 & 0.61 \\
            \midrule
            \multirow{4}{*}{Randomized AR} & RandAR-L~\cite{pang2025randar} & 343M & 2.55 & 288.8 & 0.81 & 0.58 & 2.66 & 58.8 & 0.78 & 0.61 \\
            & RandAR-XL~\cite{pang2025randar} & 775M & 2.25 & 314.2 & 0.80 & 0.60 & 2.58 & 60.3 & 0.77 & 0.62 \\
            \cmidrule(lr){2-11}
            & \cellcolor{gray!20}\method{}-L & \cellcolor{gray!20}461M & \cellcolor{gray!20}1.98 & \cellcolor{gray!20}322.0 & \cellcolor{gray!20}0.82 & \cellcolor{gray!20}0.58 & \cellcolor{gray!20}2.20 & \cellcolor{gray!20}110.1 & \cellcolor{gray!20}0.82 & \cellcolor{gray!20}0.58 \\
            & \cellcolor{gray!20}\method{}-XL & \cellcolor{gray!20}955M & \cellcolor{gray!20}1.65 & \cellcolor{gray!20}333.2 & \cellcolor{gray!20}0.80 & \cellcolor{gray!20}0.62 & \cellcolor{gray!20}2.07 & \cellcolor{gray!20}111.3 & \cellcolor{gray!20}0.81 & \cellcolor{gray!20}0.60 \\
            \bottomrule
        \end{tabular}
    }
    \label{tab:baseline}
\end{table*}
\begin{figure}[t]
  \centering
  
  \vspace{-12pt}
  \includegraphics[width=\textwidth]{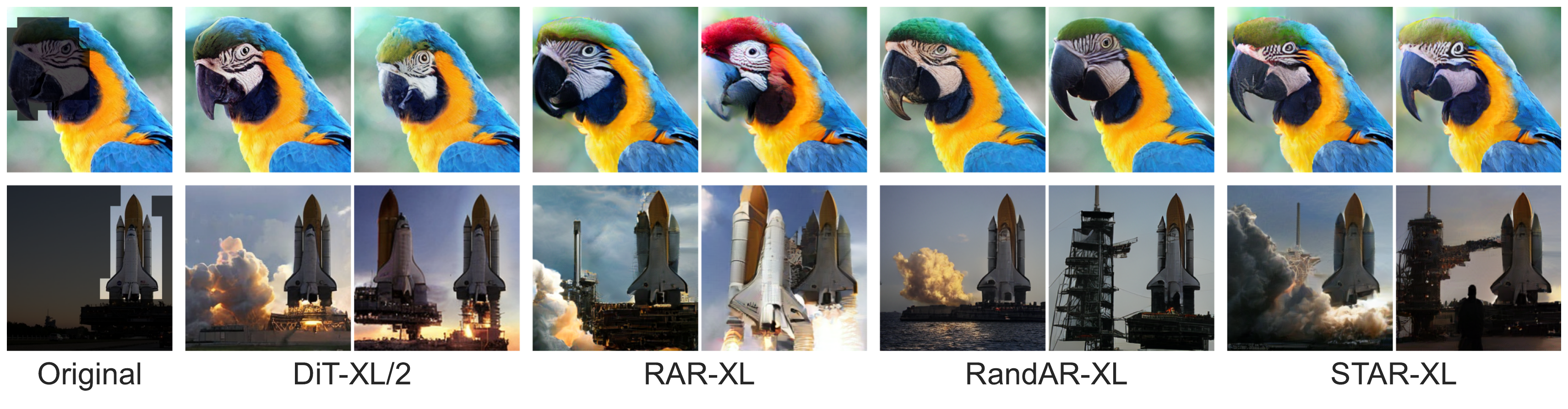}
  
  \vspace{-10pt}
  \caption{Qualitative image inpainting examples of class-conditional generative models.}
  
  \vspace{-14pt}
  \label{fig:inpainting_qualitative}
\end{figure}

\subsubsection{Comparison with the Image Inpainting}
\begin{figure}[t]
  \centering
  \includegraphics[width=\textwidth]{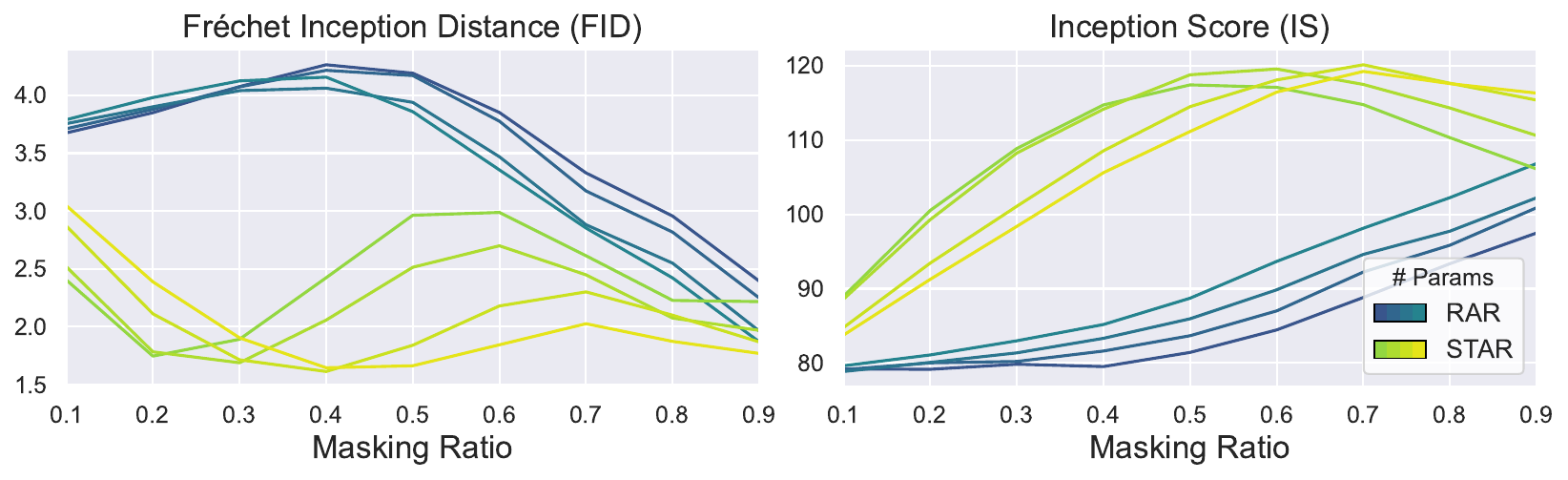}
  
  \vspace{-10pt}
  \caption{Detailed comparison of RAR~\cite{yu2025randomized} and \method{} in the image inpainting across four parameter configurations and varying masking ratios, measured by FID and IS.}
  
  \vspace{-16pt}
  \label{fig:masking}
\end{figure}

In \Cref{tab:baseline}, we further compare the image inpainting performance of \method{} with DiT, RAR, and RandAR, as well as their image generation performance. 
We provide the partial observation as a prefix for \method{} and RandAR, and through teacher forcing for RAR, and latent blending in Blended Latent Diffusion~\cite{avrahami2023blended} for DiT.
We observe that RAR and DiT exhibit a significant decrease in inpainting performance relative to their generation performance, and that some samples show low coherence with partial observations as depicted in \Cref{fig:inpainting_qualitative}.
This demonstrates that exposing bidirectional dependency between image patches only during training, or only with noisy partial observations of the reverse-time diffusion process, is insufficient to achieve native image inpainting.
Meanwhile, RandAR maintains the inpainting performance but shows inferior performance metrics compared to \method{} in both generation and inpainting.
Moreover, we find that scaling the parameter size significantly improves FID at 0.4-0.6 masking ratios, as shown in \Cref{fig:masking}.
We speculate that sufficient sample diversity, as reflected in the trend of Recall in \Cref{tab:main}, enables the incorporation of ambiguous yet informative partial observations.
Still, RAR consistently performs worse than \method{}, and its performance can be restored only at high masking ratios that resemble generation scenarios.

\subsection{Ablation Studies and Analysis}

\subsubsection{Advantage from the Locality}
\method{} utilizes the traversal orders of uniform spanning trees that form a subset of random permutations (i.e $\tau_s \big( \mathcal{T}(G, r) \big) \subset S_N$).
Given that RAR utilizes random permutation as a pretext task for the raster-scan order by annealing strategy, $S_N \setminus \tau_s \big( \mathcal{T}(G, r) \big)$ containing sequence orders that do not preserve adjacent region predicting might be similarly beneficial for sequence orders in $\tau_s \big( \mathcal{T}(G, r) \big)$.
In this sense, we compare performance changes across sequence order randomization strategies under the same computational budget as \method{}-B, as shown in \Cref{tab:strategy}.
However, we find that random permutation cannot outperform the default strategy of \method{}, whether using the traversal order of uniform spanning trees at inference time or the annealing strategy.
We suspect that sequence orders that do not preserve locality are not advantageous for convergence speed, and the lack of improvement from adopting uniform spanning trees at inference time or after the order annealing indicates that they do not serve as a meaningful pretext task beyond a single predefined sequence order.
Therefore, it demonstrates the efficiency of STAR in providing a structured random family of locality-preserving sequence orders that are sparse in random permutation, without requiring a complex annealing strategy.

\subsubsection{Advantage from the Center Bias}
\begin{figure}[t]

  \vspace{-2pt}
  \centering
      \begin{subfigure}{0.49\columnwidth}
        \includegraphics[width=\columnwidth]{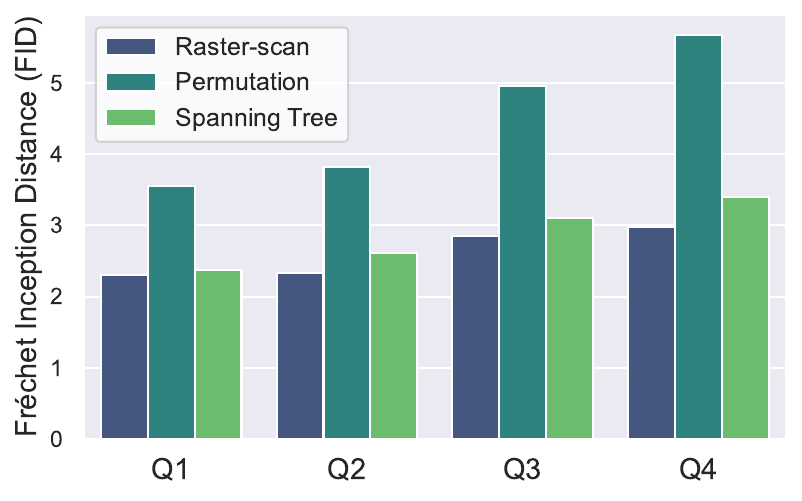}
        \caption{}
      \label{fig:quartile}
      \end{subfigure}
      \hfill
      \begin{subfigure}{0.49\columnwidth}
        \includegraphics[width=\columnwidth]{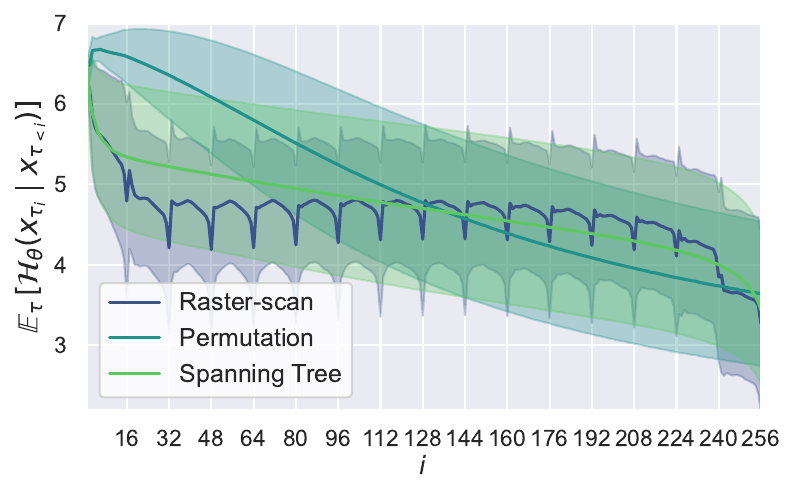}
        \caption{}
        \label{fig:sequence_entropy}
      \end{subfigure}
      
      \vspace{-6pt}
      \caption{(a) FID measured in subclasses divided according to the concentration of objects in the central region. Moving from Q1 to Q4, objects in subclasses tend to disperse farther from the center, implying a weaker center bias. (b) Conditional model entropy according to the token position in sequence orders. Sequence orders with better performance tend to have a uniform scale of conditional model entropy across token positions.}
      
      \vspace{-16pt}
\end{figure}

We select random corners as the root node for BFS traversal in \method{} from the assumption that starting from non-salient regions is advantageous under center bias, according to the observation in \Cref{sec:method_analysis} and prior works~\cite{esser2021taming, el2024scalable, yu2025randomized} that consistently report that the raster-scan order, which starts from the corner, outperforms other choice of structured sequence orders.
We further analyze this design choice by dividing ImageNet classes into four quartiles based on the strength of the center bias, measured by the standard deviation of relative object distances from the image center.
Specifically, we measure FID using 50k generated images and reference images from the entire training dataset, only for the subclasses within each quartile, as shown in \Cref{fig:quartile}.
We observe that BFS traversal of uniform spanning trees achieves performance comparable to raster-scan order at higher center bias and consistently outperforms random permutation at all levels.
We also find that performance degrades as center bias decreases regardless of sequence order choice, which is likely due to increased uncertainty about object locations resulted from reduced center bias in images.
Therefore, these results reveal that randomly selected corners as the root are well-suited to the case with strong center bias, and still generalize better than random permutation under weaker center bias.

\begin{table}[t]
    \centering
    \caption{Change in generation and inpainting performance according to the sequence order randomization strategy. "Start" and "End" denote the sequence order at the start or end, and perform order annealing at 0.5 to 0.75 of the total training steps~\cite{yu2025randomized}.}
    
    \vspace{-8pt}
    \adjustbox{max width=\textwidth}{
        \setlength{\tabcolsep}{3pt}
        \begin{tabular}{ccccccccccc}
            \toprule
            \multicolumn{2}{c}{\textbf{Training}} & \multirow{2}{*}{\textbf{Inference}} & \multicolumn{4}{c}{\textbf{Generation}} & \multicolumn{4}{c}{\textbf{Inpainting}} \\
            \cmidrule(lr){1-2}\cmidrule(lr){4-7}\cmidrule(lr){8-11}
            \footnotesize\textbf{Start} & \footnotesize\textbf{End} & & \footnotesize\textbf{FID$(\bm{\downarrow})$} & \footnotesize\textbf{IS$(\bm{\uparrow})$} & \footnotesize\textbf{P}$(\bm{\uparrow})$ & \footnotesize\textbf{R}$(\bm{\uparrow})$ & \footnotesize\textbf{FID$(\bm{\downarrow})$} & \footnotesize\textbf{IS$(\bm{\uparrow})$} & \footnotesize\textbf{P}$(\bm{\uparrow})$ & \footnotesize\textbf{R}$(\bm{\uparrow})$ \\
            \midrule
            Raster-scan & Raster-scan & Raster-scan & 2.04 & 266.5 & 0.80 & 0.59 & 3.91 & 82.7 & 0.75 & 0.60 \\
            \midrule
            Permutation & Permutation & Permutation & 3.57 & 291.4 & 0.78 & 0.57 & 2.57 & 108.7 & 0.82 & 0.56 \\
            Permutation & Permutation & Spanning Tree & 3.58 & 234.9 & 0.76 & 0.60 & 2.43 & 105.8 & 0.83 & 0.56 \\
            Permutation & Spanning Tree & Spanning Tree & 2.31 & 285.2 & 0.81 & 0.57 & 2.35 & 107.9 & 0.83 & 0.56 \\
            \midrule
            \rowcolor{gray!20}
            Spanning Tree & Spanning Tree & Spanning Tree & 2.24 & 295.3 & 0.82 & 0.57 & 2.39 & 108.8 & 0.83 & 0.56 \\
            \bottomrule
        \end{tabular}
    }

    \vspace{-10pt}
    \label{tab:strategy}
\end{table}

\subsubsection{Difference in the Conditional Model Entropy}
We have verified that prior knowledge observed in the image is still preserved in the token representations in \Cref{sec:method_analysis}, but it does not fully capture differences in the characteristics of sequence orders.
In this sense, we compare the conditional model entropy of models trained with raster-scan order, random permutation, and BFS traversal of the uniform spanning tree, based on the token position in the linearized sequence.
In \Cref{fig:sequence_entropy}, we observe that the conditional model entropy of the AR model trained on the BFS traversal of the uniform spanning tree exhibits characteristics similar to those of the AR model with raster-scan order based on the scale change with token position.
Specifically, random permutation shows high conditional model entropy at the initial token positions, then decays quickly after half of the token positions, unlike others, which show relatively constant scale across token positions.
Given that both cases show better performance compared to random permutation, we speculate that uniformity of the prediction difficulty yields a beneficial improvement in convergence speed, similar to the claim of AIM~\cite{el2024scalable}.

\begin{table}[t]
    \centering
    \caption{Comparison of the DFS and BFS with root selection strategies in the rejection sampling for postfix completion, measured by the number of trials and failure rate.}
    \vspace{-8pt}
    \resizebox{\textwidth}{!}{
        \setlength{\tabcolsep}{5pt}
        \begin{tabular}{cccccccccccc}
            \toprule
            \multirow{2}{*}{\textbf{Metric}} & \multirow{2}{*}{\textbf{Traversal}} & \multirow{2}{*}{\textbf{Root}} & \multicolumn{9}{c}{\textbf{Masking Ratio}} \\
            \cmidrule(lr){4-12}
            & & & \textbf{0.1} & \textbf{0.2} & \textbf{0.3} & \textbf{0.4} & \textbf{0.5} & \textbf{0.6} & \textbf{0.7} & \textbf{0.8} & \textbf{0.9} \\
            \midrule
            \multirow{4}{*}{Trial (\#)} & \multirow{2}{*}{DFS} & Random & 60.45 & 52.09 & 44.63 & 36.87 & 29.31 & 23.28 & 13.62 & 6.12 & 2.22 \\
            & & Farthest & 89.37 & 80.25 & 70.45 & 53.70 & 23.33 & 19.70 & 14.29 & 6.19 & 2.23 \\
            \cmidrule(lr){2-12}
            & \multirow{2}{*}{BFS} & Random & 5.59 & 3.56 & 2.76 & 2.11 & 1.60 & 1.49 & 1.44 & 1.28 & 1.15 \\
            & & Farthest & \cellcolor{gray!20}4.41 & \cellcolor{gray!20}3.03 & \cellcolor{gray!20}2.50 & \cellcolor{gray!20}1.95 & \cellcolor{gray!20}1.57 & \cellcolor{gray!20}1.51 & \cellcolor{gray!20}1.43 & \cellcolor{gray!20}1.27 & \cellcolor{gray!20}1.15 \\
            \midrule
            \multirow{4}{*}{Failure (\%)} & \multirow{2}{*}{DFS} & Random & 50.1 & 41.8 & 33.0 & 24.9 & 17.4 & 11.6 & 3.0 & 0.0 & 0.0 \\
            & & Farthest & 82.3 & 70.0 & 57.9 & 40.4 & 12.3 & 9.8 & 3.6 & 0.0 & 0.0 \\
            \cmidrule(lr){2-12}
            & \multirow{2}{*}{BFS} & Random & 0.0 & 0.0 & 0.0 & 0.0 & 0.0 & 0.0 & 0.0 & 0.0 & 0.0 \\
            & & Farthest & \cellcolor{gray!20}0.0 & \cellcolor{gray!20}0.0 & \cellcolor{gray!20}0.0 & \cellcolor{gray!20}0.0 & \cellcolor{gray!20}0.0 & \cellcolor{gray!20}0.0 & \cellcolor{gray!20}0.0 & \cellcolor{gray!20}0.0 & \cellcolor{gray!20}0.0 \\
            \bottomrule
        \end{tabular}
    }
    \vspace{-14pt}
    \label{tab:trial}
\end{table}

\subsubsection{Overhead from the Spanning Tree}
\method{} incurs additional overhead from sampling a uniform spanning tree and performing BFS on each instance, both in training and in inference.
Still, we find that $0.56$ ms is incurred per instance on average, corresponding to $< 0.0004\%$ of the total inference time across four STAR configurations, which is sufficiently small to be negligible.
However, rejection sampling for postfix completion can induce multiple trials of uniform spanning tree sampling, so we check the efficiency of rejection sampling depending on the choice of DFS and BFS with the farthest root selection strategy, as shown in \Cref{tab:trial}.
We measure the number of trials required to sample a spanning tree that satisfies the condition for postfix completion under the corresponding traversal order, for 50k random connected masks with masking ratios ranging from 0.1 to 0.9 in increments of 0.1. 
The maximum number of trials for each random mask is bounded at 100, and cases that exceed this bound are considered failures.
We find that DFS requires significantly more trials than BFS and fails more frequently, demonstrating the validity of selecting BFS as the default traversal strategy in \method{}.
In particular, we find that we can sample the spanning tree with fewer than 5 trials using BFS with the farthest root selection strategy, which reduces the number of trials at relatively small masking ratios.

\section{Discussion}
\label{sec:discussion}

\begin{figure}[t]
  \centering

  \includegraphics[width=\textwidth]{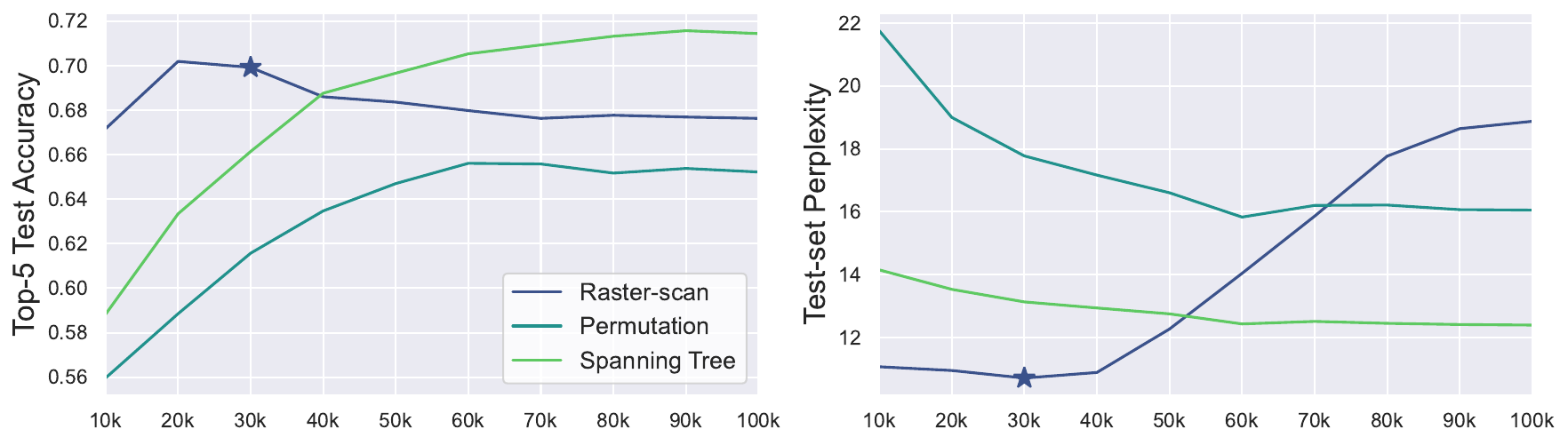}
  
  \vspace{-8pt}
  \caption{Comparison of sequence orders in representation learning. We mark the point where validation loss starts to increase, which is observed only for the raster-scan order.}
  
  \vspace{-14pt}
  \label{fig:cifar100}
\end{figure}

We investigate \method{} primarily with a focus on visual generation, but AR modeling also has another research direction: representation learning~\cite{chen2020generative, el2024scalable}.
In this context, we train models with the same configuration as iGPT-S~\cite{chen2020generative} using only the CIFAR-100 dataset~\cite{krizhevsky2009learning} for 100k steps, while changing the sequence order.
In \Cref{fig:cifar100}, we measure the top-5 test accuracy from the linear probe using mean-pooled intermediate hidden states at the 2/3 point, along with test-set perplexity, for checkpoints saved at 10k step intervals.
We confirm that uniform spanning trees achieve superior linear probe results while maintaining robustness against overfitting compared to raster-scan order, and achieve reasonable convergence speed compared to random permutation.
This suggests that future work on \method{} can be investigated within the context of representation learning.

\section{Conclusion}
\label{sec:conclusion}

In this work, we propose \methodfull{} (\method{}) modeling, which adopts the BFS traversal order of the uniform spanning tree rooted at the corner of the image patch lattice as the sequence order randomization strategy for AR visual generation models.
It structurally reflects prior knowledge of images, thereby maintaining generation performance without a complex training strategy, while still supporting flexible sequence orders to achieve region-level modification, which were the weaknesses and strengths of permutation AR modeling.
Notably, a remarkable feature of \method{} is its simplicity, which requires only minimal changes to the widely adopted conventional AR modeling architecture to support both robust image generation and image inpainting.
We hope this simplicity of \method{} can serve as a basic building block for future AR visual modeling work and be extended to early-fusion multimodal generative models.
\section*{Acknowledgements}

This work was supported by LG AI Research. 
This work was supported by the Korea Institute of Science and Technology (KIST) Institutional Program (Project No. 26E0061).
This work was partly supported by the Institute of Information \& communications Technology Planning \& Evaluation (IITP) grant funded by the Korean Government (MSIT) (No. RS-2021-II211343, Artificial Intelligence Graduate School Program (Seoul National University)), the Technology Innovation Program (RS-2025-25456760, Development of a humanoid robot specialized in chemical processes based on AI foundation model) funded By the Ministry of Trade, Industry and Resources (MOTIR, Korea). 
We express special thanks to Korea Association for AI \& ICT Promotion (KAIT) GPU project.
The ICT at Seoul National University provides research facilities for this study.

\clearpage
\bibliographystyle{splncs04}
\bibliography{main}

\clearpage
\title{Supplementary Material for\\Spanning Tree Autoregressive Visual Generation}

\titlerunning{Spanning Tree Autoregressive Visual Generation}

\author{
    Sangkyu Lee\inst{1,2}$^ ,$\thanks{Work mainly done during internship at LG AI Research.}\orcidlink{0009-0000-8009-1990} \and Changho Lee\inst{3}\orcidlink{0009-0007-7112-910X} \and Janghoon Han\inst{3}\orcidlink{0009-0007-3937-1764} \and Hosung Song\inst{3}\orcidlink{0009-0007-4044-4481} \and \\
    Tackgeun You\inst{2}\orcidlink{0009-0009-4253-9139} \and Hwasup Lim\inst{2}\orcidlink{0000-0003-2957-668X} \and Stanley Jungkyu Choi\inst{3}\orcidlink{0009-0001-0297-4269} \and \\
    Honglak Lee\inst{3,4}\orcidlink{0000-0002-4109-327X} \and Youngjae Yu\inst{5}$^{,\dagger}$\orcidlink{0000-0002-5867-0782}
}

\authorrunning{Lee et al.}

\institute{
    Yonsei University, Seoul, Korea \and 
    Korea Institute of Science and Technology, Seoul, Korea \and
    LG AI Research, Seoul, Korea \and
    University of Michigan, Ann Arbor, USA \and
    Seoul National University, Seoul, Korea
}

\maketitle
\begingroup
\renewcommand{\thefootnote}{$\dagger$}
\footnotetext{Corresponding author}
\endgroup

\appendix

\renewcommand{\thefigure}{\Alph{figure}}
\renewcommand{\thetable}{\Alph{table}}
\renewcommand{\thealgorithm}{\Alph{algorithm}}

\renewcommand{\theHfigure}{\Alph{figure}}
\renewcommand{\theHtable}{\Alph{table}}
\renewcommand{\theHalgorithm}{\Alph{algorithm}}

\setcounter{figure}{0}
\setcounter{table}{0}
\setcounter{algorithm}{0}

\section{Further Discussion about Violations of the Assumption}
\label{app:multiturn}

\begin{figure}[h]
  \centering
  
  \vspace{-6pt}
  \includegraphics[width=\textwidth]{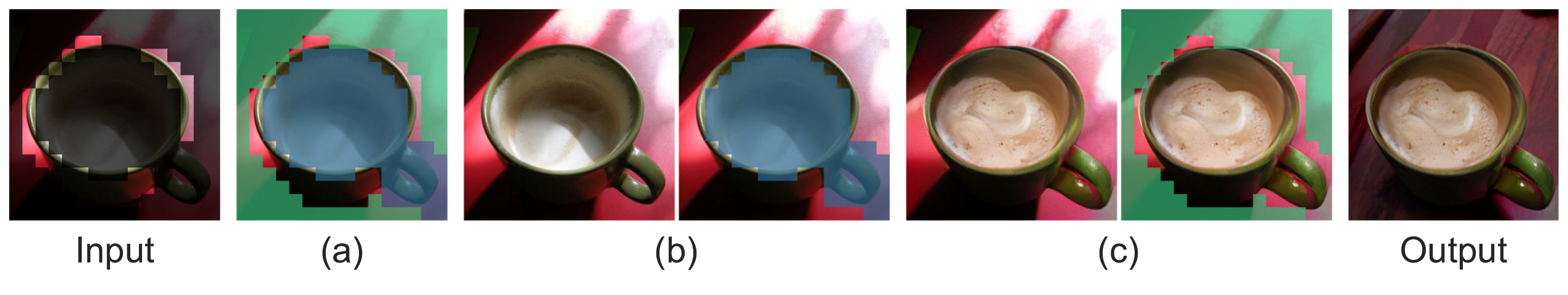}
  
  \vspace{-10pt}
  \caption{Example of multi-turn postfix completion inference for image inpainting in violation of the assumption about masking. We can divide the entire mask that covers all corners and divides partial observations into disconnected graphs, as shown in (a), and then perform postfix completion for each mask gradually, as shown in (b) and (c).}

  \label{fig:multiturn}
\end{figure}

Although violations of the assumption that the entire mask $M$ does not cover all corners and does not divide partial observations into disconnected graphs are relatively rare across the entire sample space of $M$, they are clearly realistic and thus constitute a limitation worth discussing.
However, this is a limitation under the constraint that we must achieve region-level modification through single-turn inference, and it can be resolved if we are allowed to perform multi-turn inference by decomposing the given $M$.
Specifically, we can decompose $M$ that violates the condition into tractable masks to gradually apply multiple postfix completions to achieve image inpainting, as depicted in \Cref{fig:multiturn}.
Therefore, we believe these scenarios need to be examined in multi-turn settings, particularly considering image editing scenarios that cannot be addressed by region-level modification.
We expect future work to elaborate on a multi-turn inference pipeline that can be combined with existing research on complex image editing scenarios, as well as to improve traversal strategies and tie-breaking rules, which are not widely explored in our work, to enable further efficient integration with previous work.

\section{Additional Implementation Details}
\label{app:implementation_details}

In \Cref{tab:config}, we enumerate detailed hyperparameters of four configurations of \method{}.
We train each model on the \texttt{a2-megagpu-16g} instance of Google Cloud Platform, resulting in 65, 80, 160, and 385 A100 GPU days across configurations per experiment.
We adopt a random masking strategy during quantitative evaluations of image inpainting to accommodate the various masking types that can occur in general scenarios.
We incrementally extend a random adjacent patch of the current mask, starting from a random patch until the given masking ratio $r$, ensuring that the mask does not cover all corners and does not divide the partial observation.
\Cref{alg:random_masking} describes the adopted random masking strategy.

\begin{table}[h]
    \centering
    
    \vspace{-12pt}
    \caption{Hyperparameter configurations of \method{} on ImageNet-1k at $256 \times 256$.}
    
    \vspace{-8pt}
    \adjustbox{max width=\columnwidth}{
        \setlength{\tabcolsep}{3pt}
        \begin{tabular}{lcccc}
            \toprule
            \textbf{Hyperparameter} & \textbf{\method{}-B} & \textbf{\method{}-L} & \textbf{\method{}-XL} & \textbf{\method{}-XXL} \\
            \midrule
            Depth & 24 & 24 & 32 & 40 \\
            Width & 768 & 1024 & 1280 & 1480 \\
            Feedforward Dim & 3072 & 4096 & 5120 & 6144 \\
            Attention Heads & 16 & 16 & 16 & 16 \\
            \midrule
            Optimizer & \multicolumn{4}{c}{AdamW} \\
            $(\beta_0, \beta_1)$ & \multicolumn{4}{c}{(0.9, 0.96)}\\
            Weight Decay & \multicolumn{4}{c}{0.03} \\
            LR Scheduler & \multicolumn{4}{c}{cosine with linear warmup}\\
            (Peak, Decay) Learning Rate & \multicolumn{4}{c}{(4e-4, 1e-5)} \\
            Warmup Ratio & \multicolumn{4}{c}{0.25} \\
            Batch Size & \multicolumn{4}{c}{2048} \\
            Training Steps & \multicolumn{4}{c}{250,000} \\
            Precision & \multicolumn{4}{c}{bfloat16} \\
            Gradient Clipping & \multicolumn{4}{c}{1.0} \\
            (Attention, Feedforward) Dropout & \multicolumn{4}{c}{(0.1, 0.1)} \\
            \midrule
            CFG Scheduler & \multicolumn{4}{c}{pow-cosine} \\
            Class Label Dropout & \multicolumn{4}{c}{0.1} \\
            Guidance (Scale, Power) & (16.0, 2.75) & (14.5, 2.55) & (6.5, 1.5) & (5.4, 1.25) \\
            Sampling Temperature & 0.98 & 0.98 & 0.98 & 0.98 \\
            \bottomrule
        \end{tabular}
    }
    \label{tab:config}
    
    \vspace{-28pt}
\end{table}
\begin{algorithm}[h]
    \caption{Random Masking Strategy for Image Inpainting}
    \label{alg:random_masking}
    \begin{algorithmic}[1]
        \REQUIRE lattice $G \coloneqq (V, E)$, ratio $r$, corners $R$, initial mask $V_M \coloneqq \{ v_M \sim V \} $
        \REPEAT
            \WHILE{$|V_M| < r \times |V|$}
                \STATE $ V_M \gets V_M \cup \big\{ v_M \sim \{v\in V \setminus V_M \mid \exists\,u\in V_M \,\, \text{s.t. } (u, v) \in E \} \big\} $
            \ENDWHILE
            \STATE $ M \gets (V_M, \{ (u, v) \mid \forall \,u,v \in V_M \,\, \text{s.t. } (u, v) \in E \}) $
        \UNTIL{$G \setminus M$ is connected, $R \not \subset M$}
        \RETURN $\text{mask } M$
    \end{algorithmic}
\end{algorithm}
\vspace{-26pt}

\section{Additional Examples of Sampled Results}

\clearpage
\begin{figure}[t]
  \centering
  
  \vspace{-8pt}
  \includegraphics[width=\textwidth]{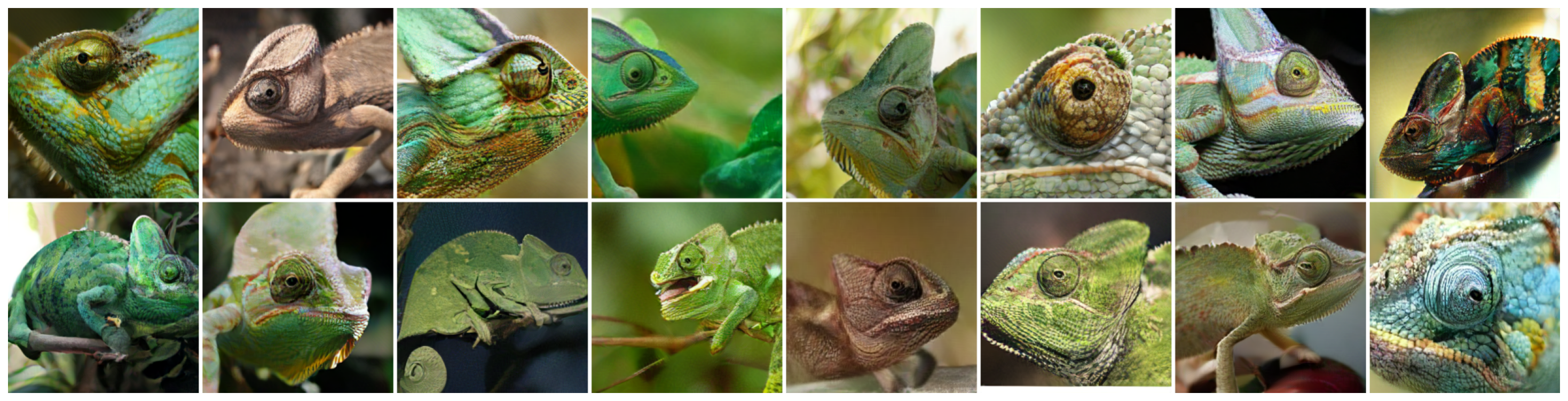}
  
  \vspace{-8pt}
  \caption{Class-conditional generation of \method{}-XXL (class ID: 47, chamaeleon).}
  
  \vspace{-10pt}
  \label{fig:47_qualitative}
\end{figure}
\begin{figure}[t]
  \centering
  
  \vspace{-8pt}
  \includegraphics[width=\textwidth]{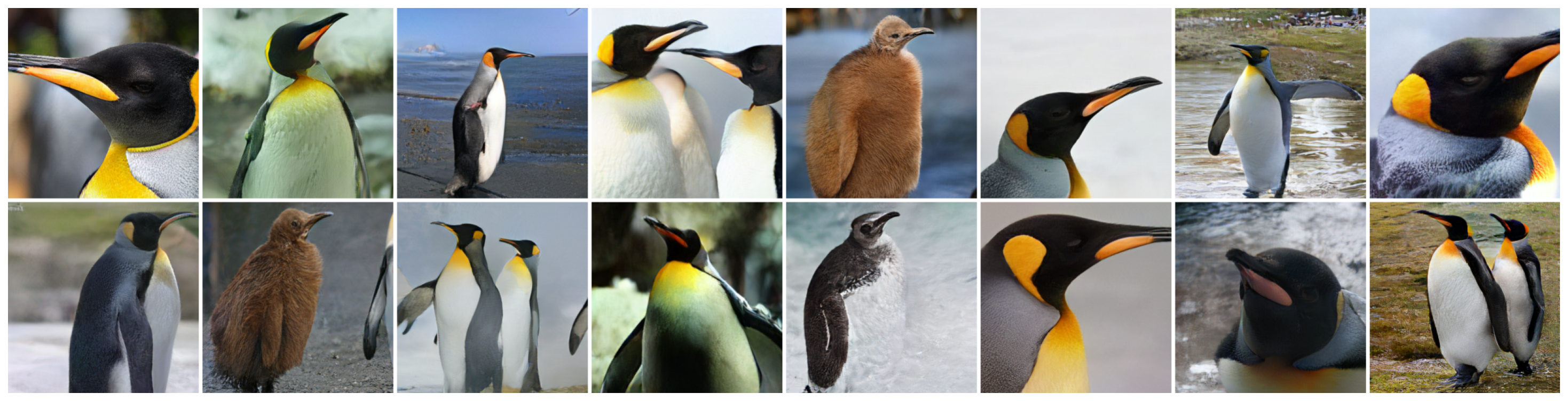}
  
  \vspace{-8pt}
  \caption{Class-conditional generation of \method{}-XXL (class ID: 145, king penguin).}
  
  \vspace{-10pt}
  \label{fig:145_qualitative}
\end{figure}
\begin{figure}[t]
  \centering
  
  \vspace{-8pt}
  \includegraphics[width=\textwidth]{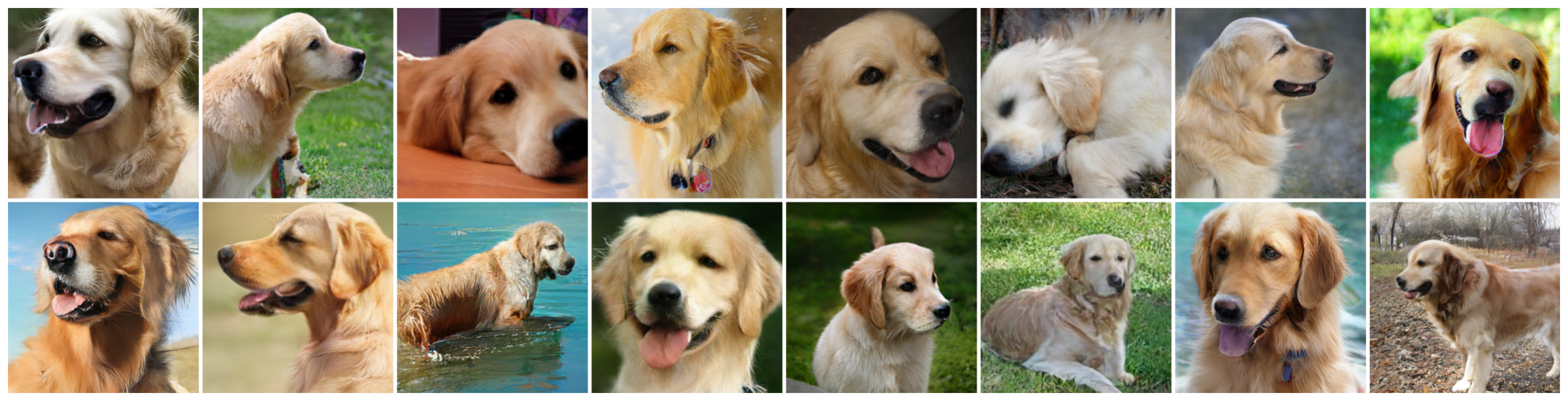}
  
  \vspace{-8pt}
  \caption{Class-conditional generation of \method{}-XXL (class ID: 207, golden retriever).}
  
  \vspace{-10pt}
  \label{fig:207_qualitative}
\end{figure}
\begin{figure}[t]
  \centering
  
  \vspace{-8pt}
  \includegraphics[width=\textwidth]{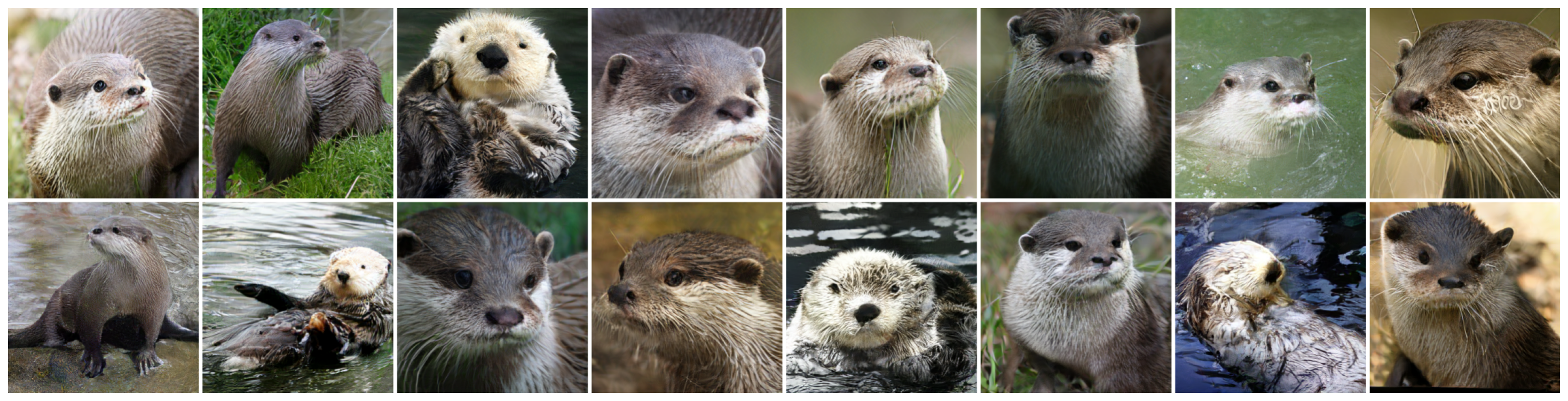}
  
  \vspace{-8pt}
  \caption{Class-conditional generation of \method{}-XXL (class ID: 360, otter).}
  
  \label{fig:360_qualitative}
\end{figure}

\clearpage
\begin{figure}[t]
  \centering
  
  \vspace{-8pt}
  \includegraphics[width=\textwidth]{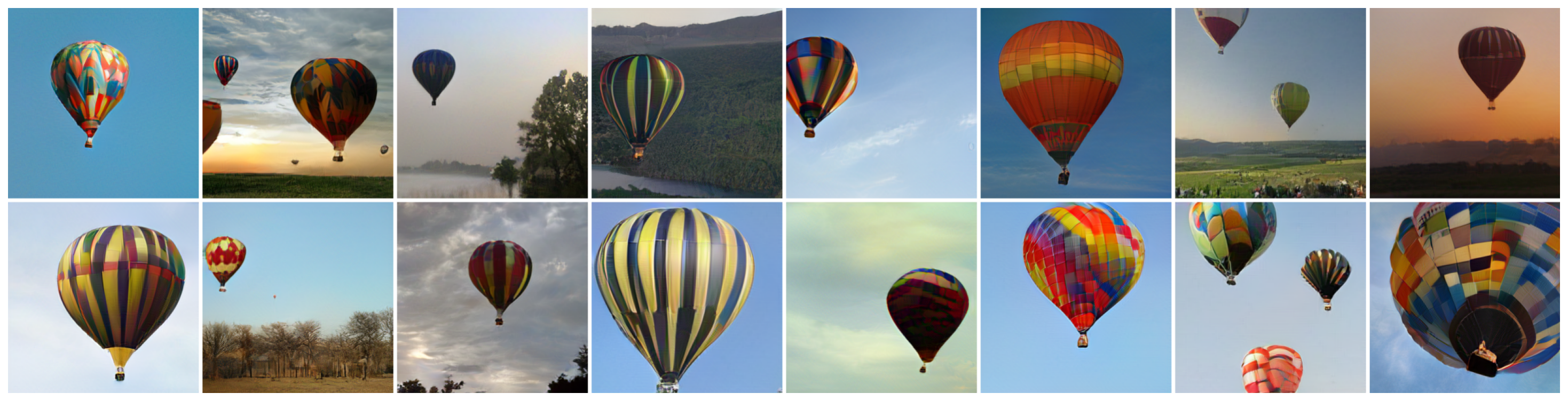}
  
  \vspace{-8pt}
  \caption{Class-conditional generation of \method{}-XXL (class ID: 417, balloon).}
  
  \vspace{-10pt}
  \label{fig:417_qualitative}
\end{figure}
\begin{figure}[t]
  \centering
  
  \vspace{-8pt}
  \includegraphics[width=\textwidth]{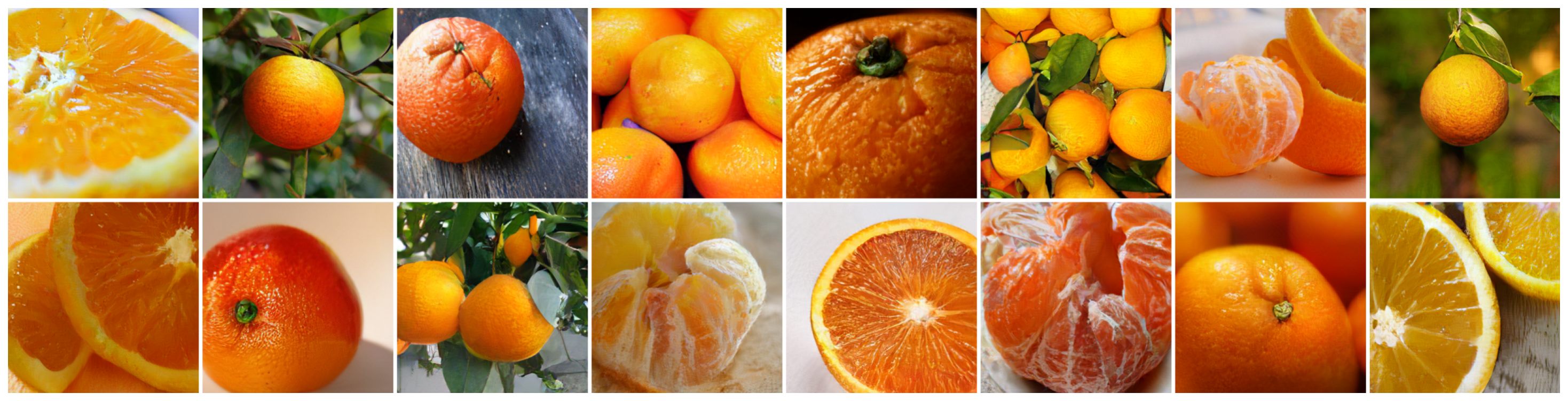}
  
  \vspace{-8pt}
  \caption{Class-conditional generation of \method{}-XXL (class ID: 950, orange).}
  
  \vspace{-10pt}
  \label{fig:950_qualitative}
\end{figure}
\begin{figure*}[t]
  \centering
  
  \vspace{-8pt}
  \includegraphics[width=\textwidth]{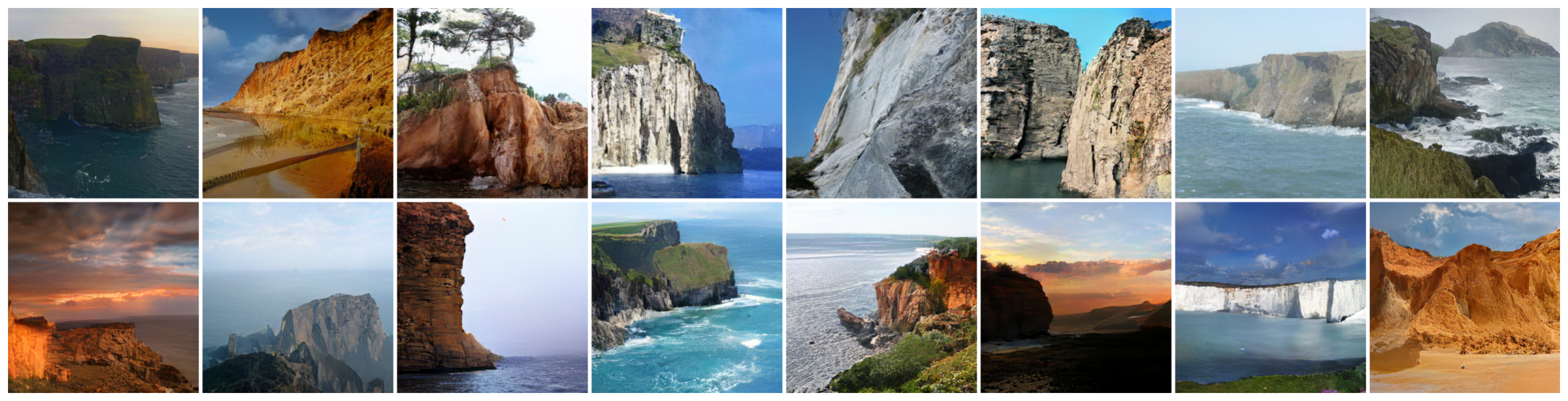}
  
  \vspace{-8pt}
  \caption{Class-conditional generation of \method{}-XXL (class ID: 972, cliff).}
  
  \vspace{-10pt}
  \label{fig:972_qualitative}
\end{figure*}
\begin{figure}[t]
  \centering
  \vspace{-8pt}
  
  \includegraphics[width=\textwidth]{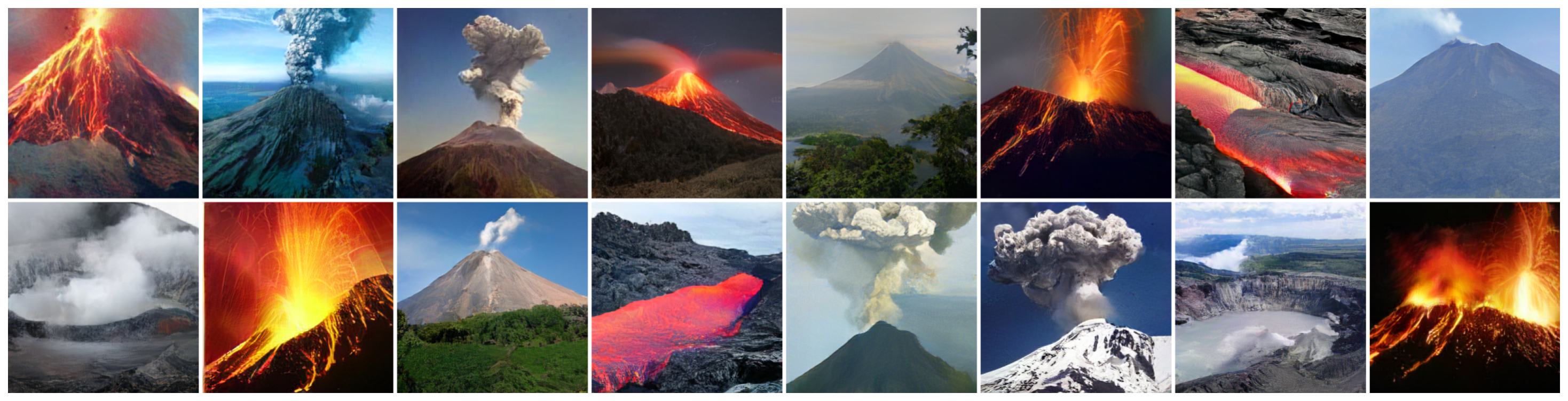}
  
  \vspace{-8pt}
  \caption{Class-conditional generation of \method{}-XXL (class ID: 980, volcano).}

  \label{fig:980_qualitative}
\end{figure}

\end{document}